\documentclass{article}
\usepackage{amsmath,amssymb,amsfonts}
\usepackage{algorithmic}
\usepackage{graphicx}
\usepackage{textcomp}
\usepackage{xcolor}
\def\BibTeX{{\rm B\kern-.05em{\sc i\kern-.025em b}\kern-.08em
    T\kern-.1667em\lower.7ex\hbox{E}\kern-.125emX}}
\usepackage{acronym} 
\acrodef{MTTS}{mixed-type time series}
\acrodef{LSTM}{long short-term memory}
\acrodef{MMSS}{Multi-modal sharable and specific feature learning}

\usepackage{multirow} 
\usepackage{booktabs} 
\usepackage{multicol} 
\usepackage {amsmath} 
\usepackage{subcaption} 
\usepackage{float} 
\usepackage{svg} 

\usepackage[backend=biber, sorting=none]{biblatex} 
\date{} 
\begin{document}

\title{\textbf{How Intermodal Interaction Affects the Performance of Deep Multimodal Fusion for Mixed-Type Time Series}\vspace{0.5cm}}

\author{%
\centering
\makebox[1\linewidth]{Simon Dietz, Thomas Altstidl, Dario Zanca, Björn Eskofier, An Nguyen}
\and
\makebox[1\linewidth]{\textit{FAU Erlangen-Nürnberg, Germany}}
\and
\makebox[1\linewidth]{\{simon.j.dietz, thomas.r.altstidl, dario.zanca, bjoern.eskofier, an.nguyen\}@fau.de}}

\maketitle

\begin{abstract}
\Ac{MTTS} is a bimodal data type that is common in many domains, such as healthcare, finance, environmental monitoring, and social media. It consists of regularly sampled continuous time series and irregularly sampled categorical event sequences.
The integration of both modalities through multimodal fusion is a promising approach for processing MTTS.
However, the question of how to effectively fuse both modalities remains open. In this paper, we present a comprehensive evaluation of several deep multimodal fusion approaches for \ac{MTTS} forecasting. Our comparison includes three fusion types (early, intermediate, and late) and five fusion methods (concatenation, weighted mean, weighted mean with correlation, gating, and feature sharing).
We evaluate these fusion approaches on three distinct datasets, one of which was generated using a novel framework. 
This framework allows for the control of key data properties, such as the strength and direction of intermodal interactions, modality imbalance, and the degree of randomness in each modality, providing a more controlled environment for testing fusion approaches.
Our findings show that the performance of different fusion approaches can be substantially influenced by the direction and strength of intermodal interactions. The study reveals that early and intermediate fusion approaches excel at capturing fine-grained and coarse-grained cross-modal features, respectively. These findings underscore the crucial role of intermodal interactions in determining the most effective fusion strategy for \ac{MTTS} forecasting.

\end{abstract}

\section{Introduction}

\begin{figure}[!htp] 
\centering
\includegraphics[width=0.9\textwidth]{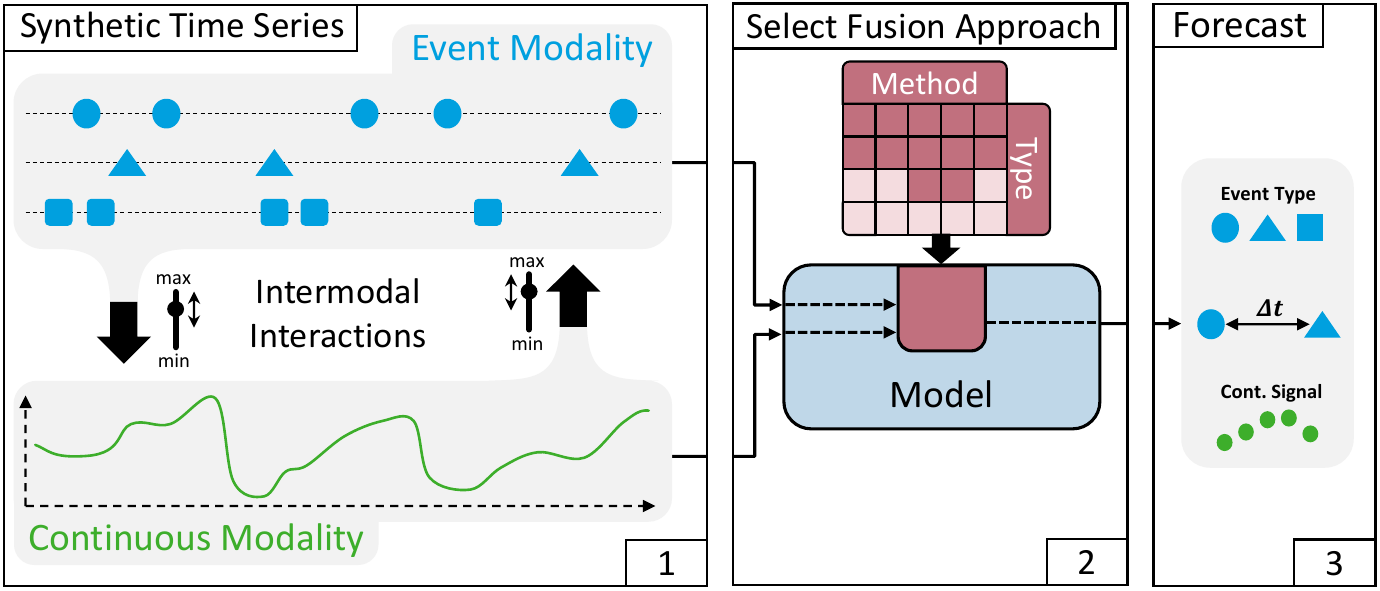}
\caption{Overview of the experimental setup. (1) Synthetic \ac{MTTS} data is generated with various interaction configurations. (2) Sequences are used to train a model for each valid combination of fusion type and method. (3) The trained models are evaluated based on their ability to forecast \ac{MTTS}. Additionally, we evaluate fusion approaches on two datasets from prior works.} 
\label{fig:visual_abstract}
\end{figure}

The analysis of multivariate time series has become a prevalent task across various domains and is now commonplace.
In this work, we address the challenge of handling \ac{MTTS} data. Specifically, we focus on the bimodal combination of regularly sampled continuous time series and irregularly sampled categorical event sequences. This data type is found in various domains, ranging from healthcare and finance to social media.
As noted, by other authors, despite its prevalence, a comprehensive comparison of how to effectively merge these two modalities is still missing \cite{debaly2023multivariate}. As a result, researchers often rely on intuition rather than rigorous evaluation when selecting a fusion strategy.
Given these circumstances, we aim to explore techniques to effectively process such data and assess the viability of different deep learning fusion approaches based on their ability to forecast \ac{MTTS}.

In the context of multimodal fusion, the fusion type refers to the commonly used differentiation between early, intermediate, and late fusion, indicating when information from sources is combined \cite{8103116}. Late fusion refers to the practice of fusing the predictions of individual networks without further feature transformations.

For this work, we term models that fuse representations of entire sequences as intermediate fusion and models that fuse partial representation as early fusion.
Traditional early fusion approaches struggle with \ac{MTTS} since there usually does not exist a matching sample from one modality for each sample of the other. Due to the irregular nature of event sequences, resampling and temporally aligning the continuous modality is challenging. Additionally, a potential imbalance in the number of samples between modalities further complicates this process.
Due to these shortcomings, we introduce an adaptation of the naive early fusion approach that temporally aligns both modalities by selectively emphasizing representations based on their temporal proximity. This allows us to apply early fusion to \ac{MTTS} without losing information or introducing noise, which is not possible with established deep learning methods.

In addition to the fusion type, we also analyze the influence different fusion methods have on a model's forecasting ability. The fusion method refers to the concrete method that is used to combine unimodal representations or predictions into multimodal ones. In this work, we compare concatenation, weighted mean, weighted mean with modality correlation, gating, and feature sharing.

We trained and evaluated each viable combination of fusion type and method on three datasets. The MazBall dataset \cite{knight2013space}, and an electrical grid dataset \cite{10.1007/978-3-030-46150-8_15}, are adapted from existing works.
In order to overcome a lack of publicly available benchmark datasets, we additionally developed a novel framework for generating synthetic \ac{MTTS} that allows for a wide range of tunable characteristics, like the direction and strength of intermodal interactions, modality imbalances and the degree of randomness in each modality. Using this framework we produced a final dataset that furthermore allowed us to analyze the effect of different interaction strengths on models forecasting ability.
\\\\
The main contributions of this work are:
\begin{itemize}
  \item An analysis of various fusion types and methods that reveals ideal fusion approaches given different data properties.
  \item A novel \ac{MTTS} generation method that allows for the configuration of various data properties.
  \item A novel early fusion method for \ac{MTTS} that overcomes the temporal alignment of modalities.
\end{itemize}
A high-level overview of the experimental setup is depicted in Figure \ref{fig:visual_abstract}.

\section{Related Work}

Multivariate time series can be found in many domains, including manufacturing, logistics, healthcare, and finance, among others \cite{YESHCHENKO2024102283}.
Some previous studies have focused on pattern-based anomaly detection \cite{feremans2020pattern}, convolution, pooling and training fusion \cite{10.1145/2663204.2663236}, and simulating \ac{MTTS} with a copula \cite{debaly2023multivariate}.
However, despite various proposed approaches, multimodal fusion remains challenging \cite{gandhi2023multimodal}.
\\\\
The taxonomy of fusion types varies, but models are commonly divided into early, intermediate, or late fusion, depending on the point at which information from modalities is fused \cite{8103116}.
In addition to the type of fusion, several different methods for fusing representations or predictions have been explored.
A straightforward approach is the concatenation of unimodal feature vectors \cite{yang-etal-2019-exploring-deep, 10.5555/3104482.3104569, Zeng_2019_CVPR}. This method allows for the fusion of representations of arbitrary size. However, as a result, the size of the feature vector is increased and might contain redundant information. 
Representations of equal size can be fused by calculating a weighted mean \cite{kumar2018multimodal}.
\Ac{MMSS} \cite{wang2015mmss} provides a compromise between concatenation and weighted mean by combining multimodal and unimodal features into one representation. This is achieved by ensuring that the learned features for different modalities share a common component while also containing modality-specific parts.
Another notable fusion approach, inspired by the \ac{LSTM} gate flow, is the gating method \cite{arevalo2017gated, narayanan2020gated}, which employs gates to emphasize input components contributing to accurate output generation selectively.
Furthermore, maximizing the correlation between unimodal representations before fusion has been shown to enhance robustness to noise \cite{yang2017deep}.

\section{Synthetic Mixed-Type Time Series Generation}
\label{sec:SMTTS generation}
Due to a lack of publicly available benchmark datasets, and in order to fully explore the effects of different data properties, we developed a novel method for generating synthetic \ac{MTTS} data with a wide range of tunable characteristics. Special consideration was given to the intermodal interaction between both modalities.
\paragraph {Continuous modality without interaction} In a scenario with no interaction between modalities, the continuous modality is simulated by an Ornstein–Uhlenbeck process \cite{uhlenbeck1930theory}.
Over time, the signal of the continuous modality gravitates towards a long-term mean $m_c$, while otherwise following a Brownian motion-like trajectory. With $\theta > 0$ as the speed of the drift towards $m_c$ and $\sigma > 0$ scaling the influence of the Wiener process $W_{t}$, the continuous signal $c(t)$ is described by the following equation:
\begin{equation}
    dc(t)=\theta (m_c(t) -x_{t})\,dt+\sigma \,dW_{t}
\end{equation}
The Euler-Maruyama method \cite{Euler-method} was used to approximate the solution to this stochastic differential equation. 
Notably, in our framework, the mean value $m_c$ is not constant but follows a predefined trajectory that is composed of several sine waves with different frequencies and amplitudes. The initial phase of this trajectory undergoes a random shift of up to one full period of the sine wave with the lowest frequency, introducing additional variability into the sequences.
\paragraph {Event modality without interaction}
Without intermodal interaction, the event sequence is generated as follows:
Given $i$ different event types ($E_1, E_2, ..., E_i$), the transition matrix $M$ is an $i \times i$ matrix where each entry $M_{jk}$ represents the probability that event type $E_k$ follows event type $E_j$. We can define a function $P(E_j|E_k)$ that gives us the conditional probability of event type $E_k$ given a preceding event type $E_j$, such that:
\begin{equation}
P(E_j | E_k) = M_{jk}
\end{equation}
Likewise, the time between event type $E_j$ occurring at time index $\tau$ and  type $E_k$ occurring at time index $\tau+1$ is given by the transition time matrix $T$ such that: 
\begin{equation}
\Delta t(E_j, E_k) = T_{jk}
\end{equation}
When there is no intermodal interaction, the two modalities of a \ac{MTTS} can be generated separately, as explained above. This way, information from one modality does not influence or improve the prediction of the other modality.
\paragraph {Modeling intermodal interactions} In order to model the interaction between both modalities, we introduce the parameters $I_{ec}\in[0,1]$ and $I_{ce}\in[0,1]$, with $I_{ec}$ and $I_{ce}$ denoting the degree of influence that the event modality has over the continuous modality and vice versa.

When there is intermodal interaction, the target mean $m(t)$ at time $t$ is the weighted mean of the default mean $m_c(t)$, which follows a static sinusoidal trajectory, and the intermodal mean $m_e(t)$, which is derived from the event sequence generated up to $t$, such that $m(t) = m_c(t) \ (1-I_{ec}) + m_e(t) \ I_{ec}$.
For the calculation of $m_e(t)$, each event type is assigned a continuous value. For this work, event type $E_0$ is assigned the value minus one and event type $E_i$ the value one. The remaining event types were assigned values uniformly distributed within the range. $m_e(t)$ is then calculated by taking the weighted average of the values associated with the past events. The weight of each event decreases exponentially with the time elapsed since the event occurred. This way, in order to predict the target mean accurately, the event sequence has to be considered, along with the past trajectory of the continuous modality.
\\\\
For modeling the interactions from the continuous to the event modality, we denote $C$, a vector that uniformly maps each unique event type $E_i$ to a value between one and minus one such that
\begin{equation}
    C=[r:r=-1+\frac{2}{i}j ,  j\in[0,1,…,i]].
\end{equation}
Based on the value of the continuous modality at time $t$, we can calculate $d=|C-c(t)|^2$ such that the sampling probability of event type $E_j$ is given by
\begin{equation}
    P(E_j|E_k, c(t)) = \text{norm}(\frac{d}{\sum d_j} \ I_{ce} + M_{jk} \ (1-I_{ce})).
\end{equation}
Finally, the term is normalized to ensure transition probabilities add up to one, and thus a proper probability distribution. By generating events this way, we ensure that the current state of the continuous modality influences the event generation.
\\\\
By adjusting the scaling factor $\sigma$ of the Brownian motion for the continuous modality and the Shannon entropy of the event transition matrix $M_{jk}$, we can regulate the degree of randomness in both modalities. The sampling rate of the continuous modality and the mean of the transition time matrix $T_{jk}$ determine how balanced the two modalities are. Using this data generation method, we can simulate a wide variety of behaviors, including modality imbalances as well as uni and bidirectional interactions.

\section{Fusion Types and Methods}

\subsection{Fusion Types}
We implemented three different fusion types, early, intermediate, and late fusion. Models can be categorized into these fusion types based on the point in an architecture, where information from both modalities is fused.

\begin{figure}[h]
\centering
\includegraphics[width=0.8\textwidth]{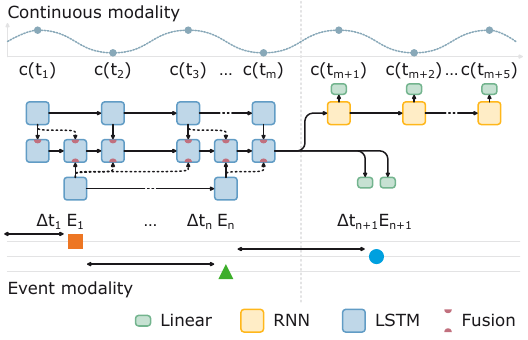}
\caption{Early fusion model architecture. The architecture consists of two unimodal \ac{LSTM} branches that process both modalities separately. Each unimodal feature representation is then fused with the latest features from the other modality at that point in the sequence. The fused features are fed into a multimodal \ac{LSTM} branch that learns the joint representation of both modalities. The final prediction is based on the output of the multimodal \ac{LSTM}.}
\label{fig:early_fusion}
\end{figure}

\begin{figure}[t]
\centering
\includegraphics[width=0.8\textwidth]{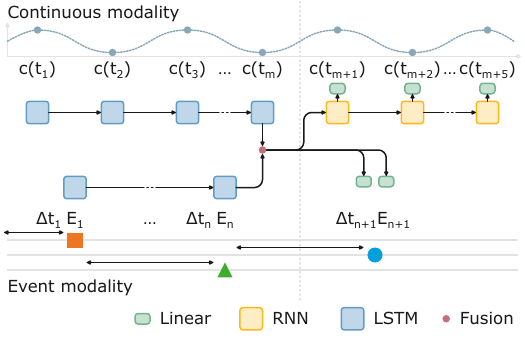}
\caption{Intermediate fusion model architecture. The architecture consists of two unimodal \ac{LSTM} branches that process the continuous and event modalities separately. The unimodal representations are then fused and passed to a prediction head that jointly generates the final forecasts. }
\label{fig:intermediate_fusion}
\end{figure}

\begin{figure}
\centering
\includegraphics[width=0.8\textwidth]{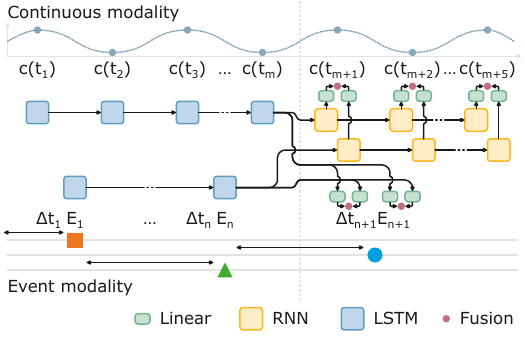}
\caption{Late fusion model architecture for \ac{MTTS}. Each unimodal branch generates one set of predictions which are then fused.}
\label{fig:late_fusion}
\end{figure}

\begin{figure}
\centering
\includegraphics[width=0.8\textwidth]{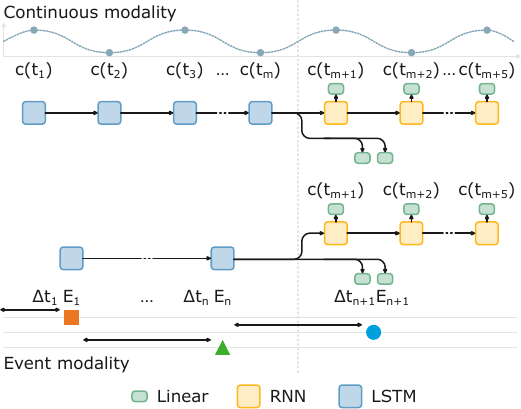}
\caption{Unimodal architectures for \ac{MTTS} forecasting. The unimodal models take one modality as input and process it using an \ac{LSTM}. The output of the \ac{LSTM} is used to generate predictions for the next continuous values, the next event type, and the time to the next event. The top figure shows the unimodal architecture for the continuous modality, and the bottom figure shows the architecture for the event modality.}
\label{fig:unimodal models}
\end{figure}

Our novel early fusion approach aims to capture the fine-grained interactions between the continuous and event modalities at each time step. We first apply an unimodal \ac{LSTM} to each modality separately to obtain unimodal features. Then, we fuse each unimodal feature representation with the latest features from the other modality at that time. The fused features are fed into a multimodal LSTM branch that learns the joint representation of both modalities, as shown in Figure \ref{fig:early_fusion}. Predictions for the next event type and the time to the next event are then each generated by a linear layer, while the continuous forecasts are generated by a recurrent neural network. 
\\\\
The intermediate fusion models extract unimodal representations from each modality using individual \ac{LSTM} branches. These representations are then fused and fed to a prediction head for further processing, as shown in Figure \ref{fig:intermediate_fusion}. 
\\\\
Our late fusion models process each modality individually and generate two sets of unimodal predictions. These predictions are then fused without further feature transformation, as shown in Figure \ref{fig:late_fusion}.
Given the inherent constraints of late fusion that do not permit further feature transformation, the selection of suitable fusion methods for this fusion type is limited to weighted mean and weighted mean with correlation.
With the late and intermediate fusion approaches, temporal modality alignment is not required since fusion occurs after each modality has been processed separately. However, in order to apply early fusion, temporal alignment of the continuous and event modalities is required.
\\\\
Additionally, we implement two baseline unimodal LSTM models that differ only in the modality used as input -- one using the continuous time series and one the event sequence. 
The final unimodal representation is then used to forecast the \ac{MTTS}. The architecture of the prediction head is identical for all models, as shown in Figure \ref{fig:unimodal models}.

\subsection{Fusion Methods}
The fusion method refers to the concrete method that is used to fuse representations or predictions. 
The fusion methods considered in this work are concatenation, weighted mean, weighted mean with modality correlation, gating, and feature sharing. 
\\\\
As a fusion method, concatenation combines unimodal feature representations $X_c$ and $X_e$ into a larger, fused representation $X_f$, such that $X_f = [X_c, X_e]$. This method is flexible and preserves the original information from each modality, but it also increases the size of the fused representation and may introduce redundant or irrelevant features. Depending on the dimension, that is being concatenated, this method also allows for the fusion of representations of different sizes. 

The weighted mean fusion method averages unimodal feature representations of equal size. $X_f = \beta X_e + (1-\beta) X_c$. For this work, the weighting factor $\beta \in [1, 0]$ was a tunable hyperparameter. In combination with late fusion, each prediction type (continuous, event type, and event time) was given a separate tunable weighting factor. 

The weighted mean fusion method with correlation works similarly, with the addition, that the correlation between modalities is calculated before fusion. In order to maximize correlation, the final loss is modified, as proposed by Yang et al. \cite{yang2017deep}, such that the final loss $L = \sum L_{\text{forecast}} - \lambda L_{\text{correlation}}$ consists of the forecasting loss $L_{\text{forecast}}$ and the correlation term $L_{\text{corr}}$ weighted with $\lambda$.

The gating fusion method uses gates similar to the ones found in \ac{LSTM} cells. From unimodal representations, we first calculate $h_e = \tanh(W_e X_e)$ and $h_c = \tanh(W_c X_c)$. 
The fused representation is then calculated as $X_f = z h_e +(1-z) h_c$, with $z=sigmoid(W_f [h_e,h_c])$. 
The matrices $W_c, W_e, W_f$ are learnable parameters. Similar gating mechanisms have been used before \cite{arevalo2017gated, narayanan2020gated}, however not for processing \ac{MTTS}.
This method allows for dynamic and adaptive fusion of unimodal feature representations, enabling the model to emphasize relevant information and ignore irrelevant or redundant features. However, it introduces additional complexity and computational cost due to the need to train the gating mechanism.

Inspired by \ac{MMSS} \cite{wang2015mmss}, feature sharing provides a compromise between the concatenation and the weighted mean fusion methods. 

The fused representation consists of two unimodal and one multimodal component. Parts of each unimodal representation are weighted and averaged, the remaining features are sliced and concatenated as follows:
\begin{equation}
    X_f = [\beta X_e[0:r] + (1-\beta) X_c[0:r], X_e[r:l_e], X_c[r:l_c]]
\end{equation} 
with $\beta \in [0, 1]$ as a weighting factor and $min(len(X_c), len(X_e)) > r > 0$ determining the fraction between unimodal and multimodal features and $l_c$ and $l_e$ denoting the length of the continuous and the event feature vector respectively. The notations $[r:l_c]$ and $[r:l_e]$ refer to the slicing of the feature vectors from the r-th to the final position.

\section{Datasets}
We use one real-world and two synthetic datasets to evaluate fusion approaches.

The MazeBall dataset was adapted from Knight et al. \cite{knight2013space}, which was captured while participants were playing a computer game. Individual button presses represent the event modality, and skin conductivity measurements captured while playing the game the continuous modality. Originally, this dataset was collected for a preference learning task.

The electrical grid dataset was adapted from Feremans et al. \cite{10.1007/978-3-030-46150-8_15}. The simulated electrical grid consists of different generators with unique sinusoidal output profiles and exponential startup and shutdown phases. The power output of simultaneously running generators is superimposed and represents the continuous modality. The startup and shutdown events of generators and additional shutdown and maintenance events represent the event modality.  Originally, the generation method was developed for anomaly detection.
Notably, the interaction between both modalities is one-sided. The event modality has a strong influence on the continuous modality, which is not the case the other way around. To increase the diversity of sequences, we added Gaussian noise with uniformly sampled standard deviations to the continuous modality and randomly sampled event probabilities.

The synthetic \ac{MTTS} dataset was generated by the method described in Section~\ref{sec:SMTTS generation}. Sequences were simulated with uni and bidirectional interaction strength parameters ranging from zero (no interaction) to one (maximum interaction), resulting in a  $200 \times 200$ grid of possible configurations. For each configuration, two sequences were generated, resulting in 80.000 training sequences. For the test set, the grid spans $20 \times 20$ possible configurations with 1000 sequences generated for each. 

\section{Experiments}
We evaluated each fusion type and method in \ac{MTTS} forecasting tasks. The models predicted the next five continuous samples, the next event type, and the time to the next event. We trained the models on a multi-objective loss function using mean squared error for the continuous signal and timestamp predictions, and cross-entropy for the event type forecast. The final loss was calculated using dynamic weight averaging \cite{liu2019end}. We used Bayesian optimization with the tree-structured Parzen estimator algorithm implemented by Optuna \cite{optuna_2019}, to tune hyperparameters such as the number of hidden units, the dropout rate, and the fusion parameters.

Our synthetic data generation allows us to set the interaction strength between modalities precisely. To investigate how different levels of intermodal interaction affect the performance of different multimodal fusion approaches, we also evaluated each model while marginalizing over the interaction strength parameters $I_{ec}$ and $I_{ce}$.

\begin{figure}[h]
\centering
\includegraphics[width=\textwidth]{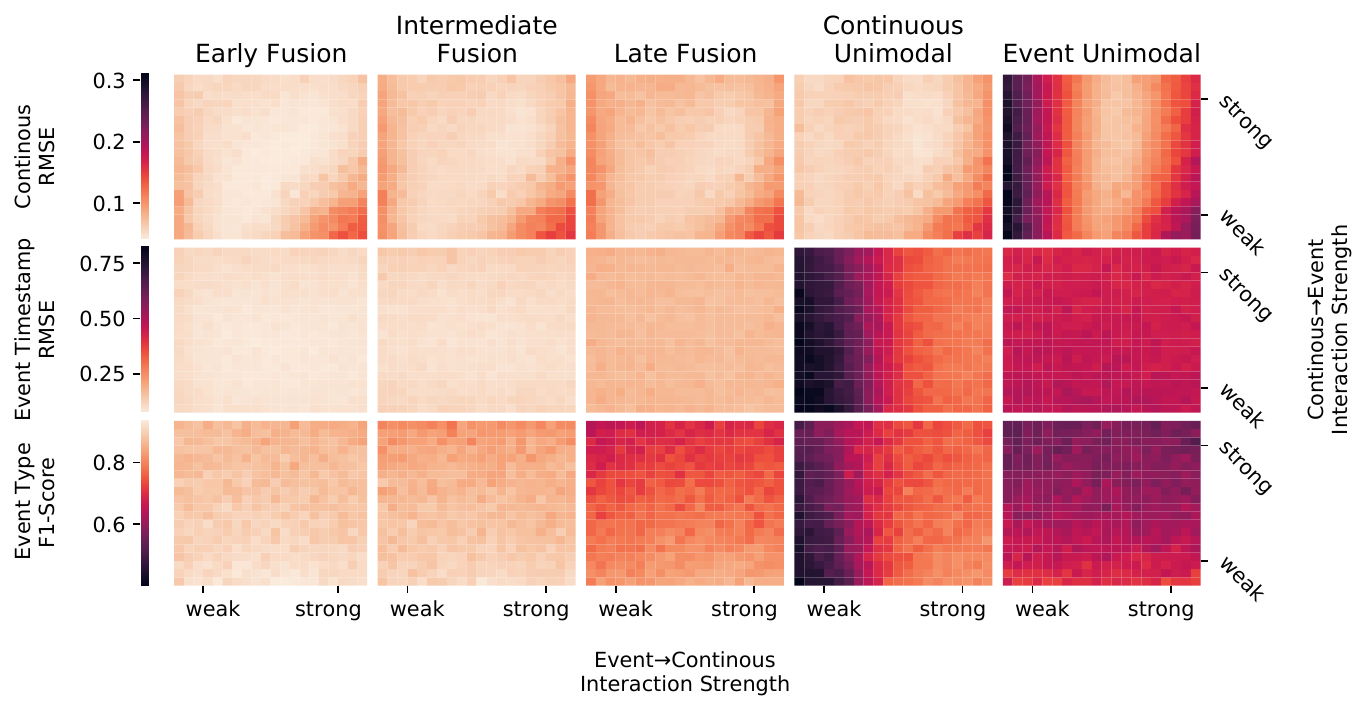}
\caption{Analysis of the effect of modality interaction strength on the forecasting ability of different multimodal fusion types (early, intermediate, late) and two unimodal baseline models receiving only the continuous or event modality as input. The figure shows the average performance of each fusion type across all interaction strength configurations in the test set. The axes represent the interaction strength of individual marginalization steps - event modality to continuous modality (Continuous→Event) and continuous to event (Event→Continuous).}
\label{fig:2d_all_types_marg}
\end{figure}

\section{Results}

\begin{table}[!htb]
\centering
\caption{Prediction metrics for different fusion types and methods on three datasets. The fusion types are early, intermediate (inter.), late, and unimodal (uni.). The fusion methods are concatenation (cat.), weighted mean (mean), weighted mean with correlation (corr.), gating, and feature sharing (share). The evaluation metrics are root mean squared error (RMSE) for the continuous modality (cont.) and the time to the next event ($\Delta t$), and $F_1$ score for the event type (event). The best, second-best, and third-best results for each dataset and metric are highlighted in bold, underlined, and italicized, respectively.}
\resizebox{\textwidth}{!}{
\begin{tabular}{l l ccc ccc ccc}
\toprule
 \multicolumn{1}{c}{} &  \multicolumn{1}{c}{} & \multicolumn{3}{c}{synthetic \ac{MTTS}} & \multicolumn{3}{c}{electrical grid}  & \multicolumn{3}{c}{MazeBall}\\
\cmidrule(lr){3-5}\cmidrule(lr){6-8}\cmidrule(lr){9-11}
type & method & cont. & $\Delta t$ & event & cont. & $\Delta t$ & event & cont. & $\Delta t$ & event \\
\midrule

\multirow{5}{*}{early}
& cat.& \textit{0.044}& \textbf{0.067}& \textbf{0.899}& 1.623& \textit{0.056}& 0.728& \underline{2.867}& \textbf{0.105}& \textbf{0.717}\\ 
& mean& \textbf{0.040}& \underline{0.082}& \textit{0.887}& 1.603& 0.057& 0.735& 3.912& \underline{0.109}& 0.702\\ 
& corr.& 0.046& 0.093& 0.887& 1.886& 0.059& 0.710& 3.348& 0.115& 0.696\\ 
& gating& \underline{0.041}& \textit{0.084}& \underline{0.887}& \textit{1.574}& \underline{0.056}& \textit{0.738}& 3.463& \textit{0.109}& 0.703\\ 
& share& 0.050& 0.099& 0.878& 1.592& 0.057& 0.737& 3.106& 0.111& 0.705\\ 
\midrule

\multirow{5}{*}{inter.}
& cat.& 0.049& 0.098& 0.876& \underline{1.573}& 0.057& \underline{0.740}& 3.747& 0.115& 0.704\\ 
& mean& 0.055& 0.111& 0.855& 1.591& 0.057& 0.731& 3.193& 0.115& 0.692\\ 
& corr.& 0.045& 0.085& 0.880& 1.738& 0.059& 0.638& 4.296& 0.112& 0.704\\ 
& gating& 0.049& 0.087& 0.872& 1.723& 0.056& 0.710& \textit{2.909}& 0.110& 0.702\\ 
& share& 0.050& 0.099& 0.873& \textbf{1.552}& \textbf{0.055}& \textbf{0.742}& 2.924& 0.113& 0.704\\ 
\midrule

\multirow{2}{*}{late}
& mean& 0.059& 0.171& 0.759& 1.751& 0.062& 0.719& \textbf{2.626}& 0.113& 0.705\\ 
& corr.& 0.058& 0.162& 0.763& 1.823& 0.062& 0.704& 3.486& 0.113& \textit{0.706}\\ 
\midrule

\multirow{2}{*}{uni.}
& cont.& 0.049& 0.472& 0.680& 1.826& 0.074& 0.579& 3.763& 0.154& 0.664\\ 
& event& 0.146& 0.445& 0.622& 2.692& 0.059& 0.737& 5.060& 0.109& \underline{0.708}\\ 
\bottomrule

\end{tabular}
}
\label{metrics_table}
\end{table}

In our study, we examined three distinct datasets: synthetic \ac{MTTS}, electrical grid data, and the MazeBall dataset. Each of these datasets presented unique characteristics and interactions between modalities, which influenced the performance of various fusion models. Table \ref{metrics_table} shows the forecasting metrics for different combinations of fusion types and methods for all three datasets.

The synthetic data was specifically configured to simulate both strong and weak unidirectional and bidirectional influences between the continuous and event modalities. In this scenario, the early fusion models, which are designed to capture low-level cross-modal features, performed the best. The top-performing fusion methods in this case were concatenation, weighted mean, and gating. These methods were able to effectively integrate the information from both modalities, leading to superior performance.

The electrical grid data, on the other hand, exhibited a unidirectional influence from the event modality to the continuous modality. This means that the events occurring within the grid had a direct impact on the power output, but not vice versa. In this case, the late fusion and unimodal models, which are less capable of capturing these types of interactions, performed poorly. The best predictions for this dataset were made by early and intermediate fusion models, which were able to better capture the unidirectional influence.

The MazeBall dataset, which is based on real-world observations, presented a different set of challenges. Good-performing models could be found across all fusion types, suggesting that the interaction between the modalities is weak. We speculate that this weak interaction limited the benefits derived from multimodality, meaning that no single fusion type outperformed any other overall.

For both the electrical grid and MazeBall datasets, we observed that unimodal models performed better when forecasting their respective input modality, but poorly when forecasting the other modality. This suggests that each modality contains unique information, which is not fully captured by the other, highlighting the importance of choosing the right model for each specific task.

The results of this study reveal that the optimal fusion strategy depends heavily on the nature of the intermodal interaction between modalities. Understanding these interactions and choosing the appropriate fusion approach is crucial for achieving the best performance in multimodal tasks.
\\\\
In order to better understand how the strength of intermodal interactions in either direction influences a models ability to forecast, we evaluated all fusion types while marginalizing over subsets of the synthetic \ac{MTTS} dataset, w.r.t. interaction strength. 
The results of this experiment are shown in Figure \ref{fig:2d_all_types_marg}. An alternative visualization can be found in Appendix \ref{appendix}.

For the baseline unimodal models, strong influences from the unseen modality generally have a weak negative effect on the performance when forecasting the input modality.
This might be a result of the unseen modality introducing noise to the input modality, which can hinder accurate forecasting
However, when these unimodal models are tasked with predicting the unseen modality itself, they benefit from medium to strong influences on the input modality.

In the case of multimodal fusion, all fusion types exhibit similar responses to different interaction configurations. When predicting the continuous modality, interactions of average intensity tend to result in the best predictions.
While the effect is much less pronounced, when predicting the time to the next event, there is a minuscule benefit from interactions of average strength for both early and intermediate fusion models.
When predicting the event modality, all fusion types appear to benefit from weak interactions from the continuous modality

In summary, depending on the forecasted modality, multimodal models often perform best, when the interaction is of medium strength. We hypothesize, that in this case, each modality carries value for forecasting itself, but the models also benefit from relevant information from the other modality.
Unimodal models are highly susceptible to interaction strength but benefit from the interaction of an unseen modality when forecasting it.

\section{Conclusion}
In this paper, we presented a novel framework for generating synthetic \ac{MTTS}, consisting of categorical event sequences and continuous time series. The framework allows for precise control over the magnitude of intermodal interaction between both modalities. 
We evaluated several fusion types (early, intermediate, and late fusion) and fusion methods (concatenation, weighted mean, weighted mean with correlation, gating, and feature sharing) on three datasets. The results indicate, that the optimal fusion strategy depends on the strength and directions of interaction between both modalities and that early and intermediate fusion approaches are capable of considering fine and coarse-grained intermodal interactions when forecasting \ac{MTTS}. 
Furthermore, we introduced a novel approach for early fusion with \ac{MTTS}, that overcomes the challenge of temporal modality alignment and performs well over all datasets.
Finally, we used our synthetic data generation method to analyze how the interaction strength influences a models ability to forecast synthetic \ac{MTTS}. The results indicate, that multimodal models often benefit from balanced interactions between modalities. Hence, our study demonstrates the importance of considering the strength of intermodal interactions when choosing a strategy for \ac{MTTS} forecasting. 
\\\\
A promising direction for future research could be the exploration of how multimodal transformer models can be adapted to \ac{MTTS}. For example, the concept of attention bottlenecks for multimodal fusion \cite{nagrani2021attention} could be applied to \ac{MTTS}. Via the number of bottleneck tokens, the degree of information sharing between modalities could be configured based on the expected modality interaction.
However, a limiting factor for research on \ac{MTTS} forecasting, remains the absence of a publicly accessible benchmark dataset.

\printbibliography

\newpage
\appendix
\counterwithin{figure}{section}
\section{Unidirectional Marginalization}
\label{appendix}
\begin{figure}[h]
\centering
\includegraphics[width=\textwidth]{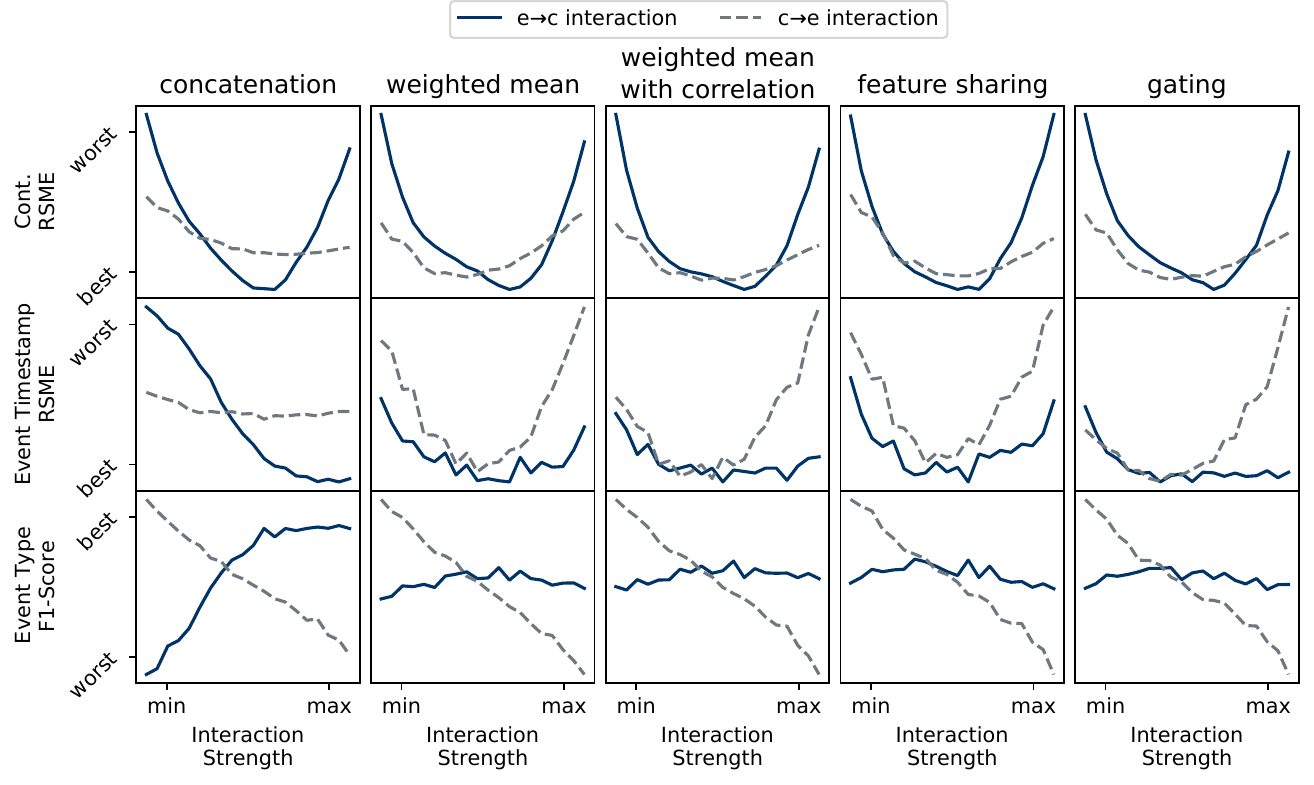}
\caption{Analysis of the effect of modality interaction strength on the forecasting ability of different multimodal fusion methods (concatenation, weighted mean, weighted mean
with correlation, feature sharing, and gating). The figure shows the average performance of each fusion type while marginalizing the interaction strength in one direction. Each line represents either interaction from the event to continuous modality (c→e) or continuous to event (e→c). The interaction scale ranges from no interaction ($I_{ce/ec} = 0$) to maximum interaction ($I_{ce/ec} = 1$).}
\label{fig:all_methods_marg}
\end{figure}

\begin{figure}[h]
\centering
\includegraphics[width=\textwidth]{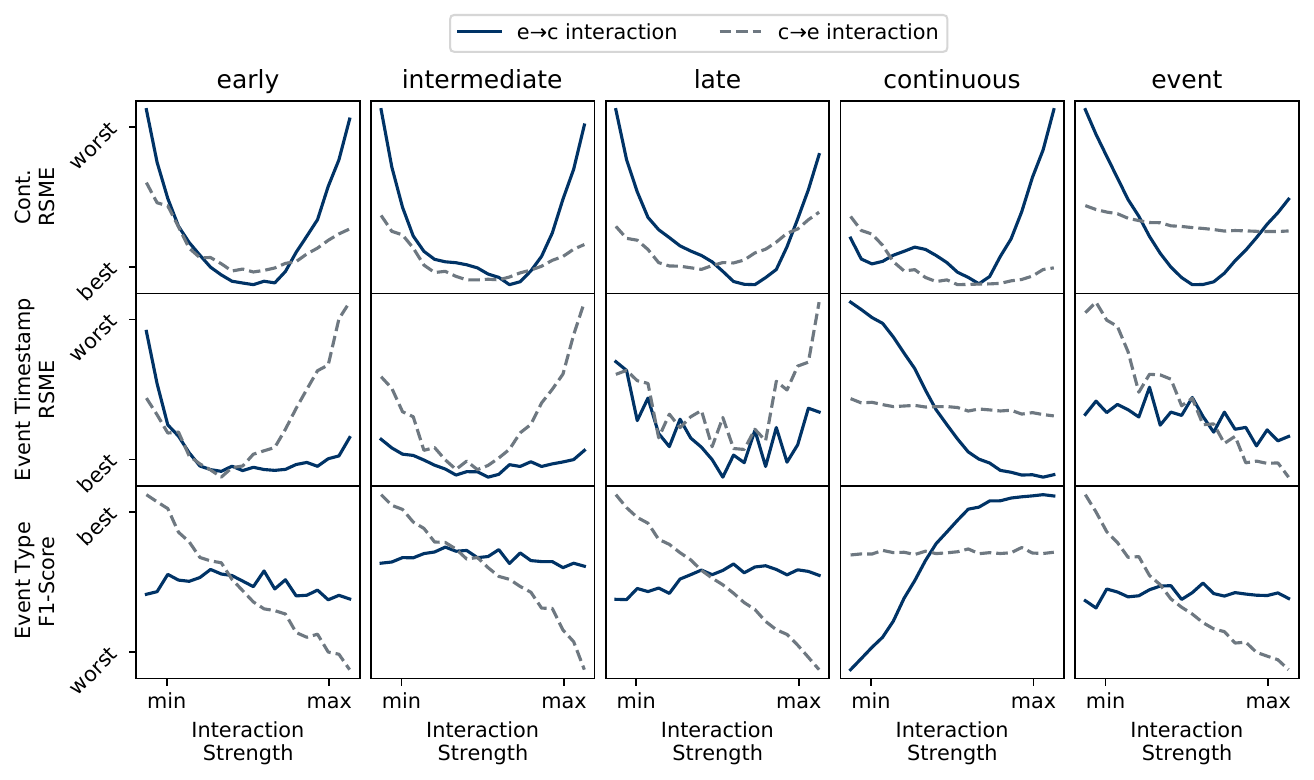}
\caption{Analysis of the effect of modality interaction strength on the forecasting ability of different multimodal fusion types (early, intermediate, late) and two unimodal baseline models receiving only the continuous or event modality as input. The figure shows the average performance of each fusion type while marginalizing the interaction strength in one direction. Each line represents either interaction from the event to continuous modality (c→e) or continuous to event (e→c). The interaction scale ranges from no interaction ($I_{ce/ec} = 0$) to maximum interaction ($I_{ce/ec} = 1$).}
\label{fig:all_types_marg}
\end{figure}

\end{document}